\pdfoutput=1
\documentclass[11pt]{article}

\usepackage{acl}
\usepackage{times}
\usepackage{latexsym}

\usepackage[T1]{fontenc}
\usepackage[utf8]{inputenc}

\usepackage{microtype}

\usepackage{inconsolata}

\usepackage[noend]{algpseudocode}
\usepackage{amsmath}
\usepackage{amsfonts}
\usepackage{amssymb}
\usepackage{enumitem}
\usepackage{multicol}
\usepackage{multirow}
\usepackage{array}
\usepackage{booktabs}
\usepackage{float}
\usepackage{mathrsfs}
\usepackage{graphicx}
\usepackage{algorithmicx}
\usepackage[ruled, lined, linesnumbered, commentsnumbered, longend]{algorithm2e}

\usepackage{amsmath,amsfonts,bm}

\def\eqref#1{equation~\ref{#1}}
\def\1{\bm{1}}

\DeclareMathAlphabet{\mathsfit}{\encodingdefault}{\sfdefault}{m}{sl}
\SetMathAlphabet{\mathsfit}{bold}{\encodingdefault}{\sfdefault}{bx}{n}

\def\gA{{\mathcal{A}}}

\def\gC{{\mathcal{C}}}

\def\gE{{\mathcal{E}}}

\def\gP{{\mathcal{P}}}
\def\gQ{{\mathcal{Q}}}
\def\gR{{\mathcal{R}}}

\def\sD{{\mathbb{D}}}
\def\sS{{\mathbb{S}}}

\newcommand\ei{$\textsc{EvIT}$}
\newcommand\eis{$\textsc{EvIT}$\ }

\newcommand\eq{event quadruple}
\newcommand\eqs{event quadruple\ }

\title{\ei: Event-Oriented Instruction Tuning for Event Reasoning}

\author{Zhengwei Tao$^{12}$ ~Xiancai Chen$^{12}$ ~Zhi Jin$^{12}$\thanks{*Corresponding authors.} ~Xiaoying Bai$^{3*}$ ~Haiyan Zhao$^{12}$ ~Yiwei Lou$^2$\\
        $^1$Key Laboratory of High Confidence Software Technologies (PKU), MOE, China\\
        $^2$School of Computer Science, Peking University\\
        $^3$Advanced Institute of Big Data \\ 
        \texttt{\{tttzw, xiancaich\}@stu.pku.edu.cn}, ~\texttt{baixy@aibd.ac.cn}\\ ~\texttt{\{zhijin, zhhy.sei\}@pku.edu.cn}
    }

\begin{document}
\maketitle
\begin{abstract}
Events refer to specific occurrences, incidents, or happenings that take place under a particular background. Event reasoning aims to infer events according to certain relations and predict future events. The cutting-edge techniques for event reasoning play a crucial role in various natural language processing applications. 
Large language models (LLMs) have made significant advancements in event reasoning owing to their wealth of knowledge and reasoning capabilities. However, smaller instruction-tuned models currently in use do not consistently demonstrate exceptional proficiency in managing these tasks. This discrepancy arises from the absence of explicit modeling of events and the interconnections of them within their instruction data.
Consequently, these models face challenges in comprehending event structures and semantics while struggling to bridge the gap between their interpretations and human understanding of events. Additionally, their limitations in grasping event relations lead to constrained event reasoning abilities to effectively deduce and incorporate pertinent event knowledge.
In this paper, we propose Event-Oriented Instruction Tuning to train our LLM named \eis specializing in event reasoning tasks. 
Specifically, we first propose a novel structure named \eqs which contains the structure and semantics of events and is complete in the event representation.
We then design event-relation learning based on the structures. 
We encapsulate the learning into the instruction-tuning formulation to better stimulate the event reasoning capacity of our model. 
We design a heuristic unsupervised method to mine \eqs from a large-scale corpus. 
At last, we finetune a Llama model on our Event-Oriented Instruction Tuning. 
We conduct extensive experiments on event reasoning tasks on several datasets. 
Automatic and human evaluations demonstrate \eis achieves competitive performances on event reasoning.
\end{abstract}

\vspace{-3mm}
\section{Introduction}
Events are instances or occurrences that form the basic semantic building units encompassing the meanings of Activities, Accomplishments, Achievements, and States~\cite{vendler1957verbs}. 
% Event semantics pertains to the interpretation of these events within textual data~\cite{tao2023eveval}. 
By employing advanced techniques and models, event reasoning aims to enable machines to comprehend the mechanism of real-world event evolution~\cite{tao2023eveval}. Under this ultimate goal, event reasoning consists of several key sub-objectives, including the understanding and reasoning about a diverse range of event inter-relations, and predicting events pertaining to certain relations. 
Reasoning events forms the foundation of sorts of NLP applications such as recommendation systems~\cite{yang2020temporal}, and question answering~\cite{souza2020event}. 

% In recent times, Large language models (LLMs) such as ChatGPT and Bloomz-175B~\cite{muennighoff2022crosslingual} have exhibited remarkable achievements in a broad spectrum of tasks. 
% These notable developments thus result in substantial advancements in this domain of event semantic processing. They have demonstrated their proficiency in understanding events, reasoning about relationships between events, and predicting future events\cite{tao2023eveval}. 
In recent times, substantial research efforts are dedicated to instructing-tuning language models to acquire the abilities for zero-shot inference such as Flan-T5 Alpaca~\cite{alpaca}, Vicuna~\cite{vicuna2023}, WizardLM~\cite{xu2023wizardlm}, and Dolly~\cite{DatabricksBlog2023DollyV2}. These models have shown the potential to enhance the language models with versatile instruction-following capabilities through fine-tuning various instruction datasets. Nonetheless, in the training of these models, the instruction-tuning data involved did not explicitly model events and their inter-relations. Consequently, these models perform inferiorly on most event reasoning tasks. The limitations observed in the instruction-tuned models stem from several fundamental factors. Firstly, these models display an inadequate understanding of event structures and semantics and show discrepancies between the model's interpretation and human comprehension of events. Secondly, the models exhibit deficiencies in comprehending the relations between events, resulting in insufficient event reasoning capabilities and the inability to effectively infer and integrate relevant event knowledge.
Based on the performances, instruction-tuning smaller language models exhibit poorer performance when contrasted with large language models (LLMs) such as ChatGPT and Bloomz-175B~\cite{muennighoff2022crosslingual}.
% which have demonstrated their proficiency in reasoning about event relations and predicting future events~\cite{tao2023eveval}. 
% These notable developments thus result in substantial advancements in this domain of event semantic processing. They have demonstrated their proficiency in understanding events, reasoning about relationships between events, and predicting future events~\cite{tao2023eveval}. 

To address these obstacles, we present \eis which is trained on our novel \textbf{Ev}ent-oriented \textbf{I}nstruction \textbf{T}uning. In our method, we incorporate explicit event modeling and event relation comprehension. Specifically, to enhance the comprehension of the structure and semantics of events, we first design a novel structure named \eq. This event-centric structure contains two events, their relation, and the background information where the fact holds. The \eqs covering contextualized events and their inter-relation knowledge would improve the model's conceptions of events. 
Based on the \eq, we develop an event-relation learning paradigm. We train \eis to predict the tail events of \eqs in both generation and discrimination manners. We further encapsulate this training process into instruction tuning with generated instruction templates. It can better stimulate the model's abilities to conduct event-related reasoning and associate event knowledge.
To implement our training, we construct \eqs from a large-scale textual corpus. We design a heuristic negative events mining algorithm to construct candidate events for discriminative event-relation training. We finetune Llama by our event-oriented instruction tuning.

We conduct extensive experiments to testify to the effectiveness of \ei. We first evaluate the performance of \eis across 8 tasks of event reasoning which are not seen during training. Among these tasks, four are held-in tasks, involving relations explicitly handled during training, while the remaining four tasks are held-out tasks. Results of automatic and human evaluations show that \eis outperforms other instruction-tuned models. 

We summarize our contributions:
\begin{itemize}[topsep=0pt]
\setlength{\parskip}{0pt}
\setlength{\parsep}{-1pt}
\setlength{\leftmargin}{-1pt}
\item[$\bullet$] We propose a novel event-oriented instruction tuning paradigm that may also shed light on other event-oriented training. We first design an event-centric structure named \eq. Based on \eq, we develop the event-relation learning. We then encapsulate the objectives into instruction-tuning.

\item[$\bullet$] We construct an event-oriented instruction-tuning dataset encompassing integrated and diversified data of events in terms of both syntax and semantics with rich relation knowledge.

\item[$\bullet$] We conduct extensive experiments on 8 datasets for testing. Results show the effectiveness of \ei.

\end{itemize}

\section{Preliminaries}
\subsection{Event Definition}
An event is something that happens involving participants~\cite{doddington2004automatic}, which may have correlations with others. Formally, let $\gE$ be an event consisting of several participants or arguments. 
% $\eva$\footnote{Here we regard the trigger as a kind of participant.}. 
Two events $\gE_u$ and $\gE_v$ can have a relation $\gR\in \sS^{\gR}$. $\sS^{\gR}$ is the universe set of event inter-relation which could cover abundant relation types such as temporality, causality, condition, prerequisites, and counterfactual~\cite{zhang2020aser}.

\subsection{Event Reasoning}
Event reasoning aims to comprehend, deduce interrelated events, or anticipate forthcoming occurrences~\cite{tao2023eveval}.  
% introduce event reasoning which is to empower machines with the capability to engage in semantic tasks guided by event-driven mechanisms.
It requires to process of queries to deduce events pertaining to specific relations~\cite{han2021ester}. These relations encompass causality, temporality, counterfactual scenarios, and intent. Distinct interconnections between events demand diverse reasoning proficiencies.

Building upon relational reasoning, the advanced objective of event reasoning revolves around predicting future events~\cite{zhao2021event}. This intricate task mandates the model to grasp events and their relations, possess substantial event-related knowledge and an understanding of event-evolution mechanisms, and ultimately integrate these aspects to prognosticate future events.

% They induce from practical needs and build the hierarchy of event semantic processing which encompasses \textsc{Understanding}, \textsc{Reasoning}, and \textsc{Prediction} of event semantics. 

% For understanding events, human demands machines to understand events correctly and consistently the same as our human beings~\cite{pedinotti2021did}. 

\section{\eis Methodology}
\subsection{Overview}
\begin{figure*}[!tb]
    \centering
    \includegraphics[width=2.1\columnwidth]{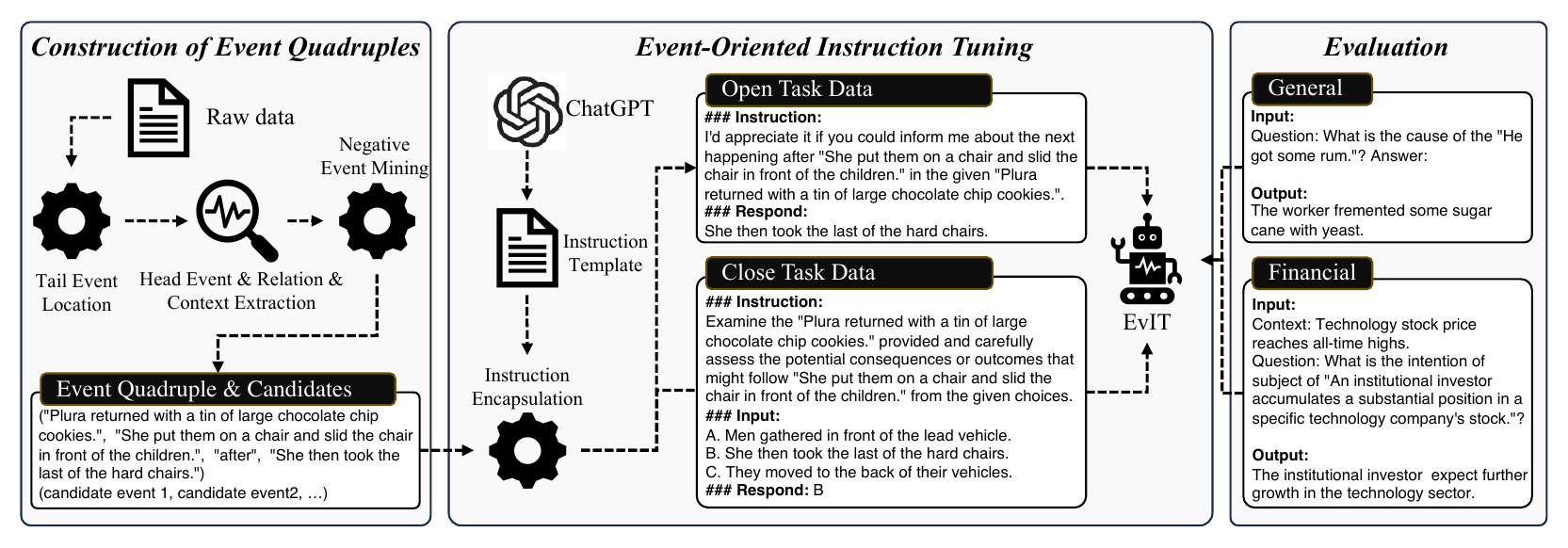}
    \caption{Overview of training process and evaluation of \ei. The training process encompasses Event-Oriented Instruction Tuning and Construction of \eq.}
    \label{fig:overview}
\end{figure*}
Our primary aim is to achieve an improved model \eis that excels in event reasoning tasks. An overview of the \eis training and evaluation process is illustrated in Figure~\ref{fig:overview}. To accomplish this objective, we begin by proposing Event-Oriented Instruction Tuning. Within this training framework, we introduce an event-centric structure denoted as \eqs along with event-relation learning. This learning is then integrated into the instruction-tuning process. Subsequently, we establish the method of construction of the \eqs and the training dataset to execute our novel training approach outlined. 

\subsection{Event-Oriented Instruction Tuning}
\label{eotp}
Large language models are first pre-trained on enormous unsupervised data and then fine-tuned on supervised data with instructions~\cite{alpaca, vicuna2023, xu2023wizardlm, DatabricksBlog2023DollyV2, openai2023gpt}. However, during all stages of training, existing LLMs are not explicitly trained to understand events and their inter-relations. This leads to several deficiencies. First, they exhibit a lack of comprehension of the structure and semantics of events.
This makes a difference between the conceptualization of these models and human understanding.
% Discrepancies prevail between the model's conceptualization and human understanding.
% Second, they exhibit deficient apprehension of relations between events. When executing event-oriented reasoning, they prove unable to adequately ascertain and integrate knowledge pertaining to the events in question. 
Second, they exhibit deficient apprehension of relations between events. When executing event reasoning, they prove unable to adequately ascertain and integrate knowledge pertaining to the events in question. This shows that these LLMs may not be able to achieve good performance in event reasoning.

% Therefore, these LLMs are not able to achieve well performance in event reasoning.

In an endeavor to mitigate these limitations, we initially introduce a novel structure referred to as \eq, which encompasses comprehensive event knowledge and their inter-relations. Subsequently, we establish event-relation learning based on this framework, ultimately encapsulating this approach into instruction tuning.

\paragraph{Event Quadruple} An \eqs $\gQ$ is:
\begin{equation}
    \gQ = (\gC, \gE^{h}, \gR, \gE^{t}),
\end{equation}
in which $\gE^{h}$ is the head event, $\gE^{t}$ is the tail event, and $\gR$ is the relation between them. $\gC$ is a paragraph of context describing the background information of both events. 
% The inter-event relation $\gR\in\sS^{\gR}$ could cover abundant event relations such as temporality, causality, condition, prerequisites, and counterfactual~\cite{zhang2020aser}. 
The \eqs $\gQ$ entails rich semantic and syntactic information of events since each $\gE$ describes an event occurring unit that aligns with human understanding. Besides, $\gQ$ is rich in event relational and structural knowledge since it precisely captures event inter-relations. 
Finally, $\gQ$ extracts the necessary information for the above events from the context. Contextual information is important for an accurate understanding of an event, because, in the absence of contextual information, the understanding of the event is prone to ambiguity.
% Lastly, $\gQ$ draws the necessary information of the mentioned events from the context. In the absence of contextual information, the comprehension of events risks ambiguity. 
In summary, using the event quadruple $\gQ$ to capture different aspects of events may reduce the risk of event misunderstanding and enhance the conceptions of structure and semantics of the events, thereby improving the accuracy of achieving event reasoning.
% In summary, the event quadruple $\gQ$ mitigates the risk and enhances the accuracy of the knowledge garnered.

\paragraph{Event-Relation Learning}
Our next objective is to leverage the \eqs to stimulate the event reasoning abilities of LLMs. The motivation is to enhance the model's understanding of event semantics, event composition, and the interpretation of event relations. We require the model learns to generate the tail event $\gE^{t}$ based on the head event $\gE^{h}$, the context $\gC$ and according to the relation $\gR$:
\begin{equation}
\label{gen}
    \gE^{t} = \mathrm{M}~(\gE^{h}, \gR, \gC).
\end{equation}
\textsc{M} is the model to be trained. Through learning to generate events, the model's comprehension of event semantics and structure was stimulated, enabling it to accomplish event reasoning tasks in a manner more aligned with human understanding. Concurrently, this process necessitated the model's apprehension of inter-event relationships, empowering it to associate pertinent event knowledge in order to conduct event relational inference. Moreover, the model learns to draw proper information from the context to answer event reasoning questions more precisely.

In order to enhance the model's event understanding capability and reduce instances of hallucination, we introduce an additional step involving multiple-choice discrimination:
\begin{equation}
\label{mc}
    \gE^{t} = \mathrm{M}~(\gE^{h}, \gR, \gC|~\sD).
\end{equation}
$\sD$ is the set of candidate events including the ground-truth tail event $\gE^{t}$ and also several negative candidates. This learning process further reinforced the model's comprehension of events and their interrelationships, enhancing the model's discriminative capabilities of event knowledge. 

\paragraph{Instruction-Tuning Encapsulation}
Incorporating event-relation learning into equations Eq.~(\ref{gen}) and (\ref{mc}) can be approached by a basic method of merging the two training procedures into generation training~\cite{tao2023unievent}. However, this approach does not successfully capture the human strategies employed in these tasks, resulting in an absence of unsupervised event reasoning abilities. In contrast, instruction-tuning techniques achieve alignment and knowledge enhancement~\cite{alpaca, vicuna2023}. Thus, we integrate event-relation learning into instruction tuning as our means to attain the desired goal.

In instruction-tuning, each dataset includes an instruction, an input, and a response. Our method involves encapsulating the input notation $\gQ$ within an instruction, adhering to a predefined template. Initially, we derive instruction templates by querying ChatGPT. Our exploration of event-relation learning encompasses $|\sS^{\gR}|$ relations, approached through two distinct formulations: generation and discrimination. Furthermore, we account for situations in which the context $\gC$ might be absent. Consequently, we require total amounts to $|\sS^{\gR}| \times 2 \times 2$ variations of instruction templates. For each kind, we ask ChatGPT to list 100 prompts with the query. We depict a query for discrimination instruction templates of $\gR=\texttt{Before}$ with context $\gC$ in Figure~\ref{fig:chatgpt} (a). More queries are in the Appendix D.
\begin{figure}[!tb]
\setlength{\belowcaptionskip}{-5mm}
    \centering
    \includegraphics[width=1\columnwidth]{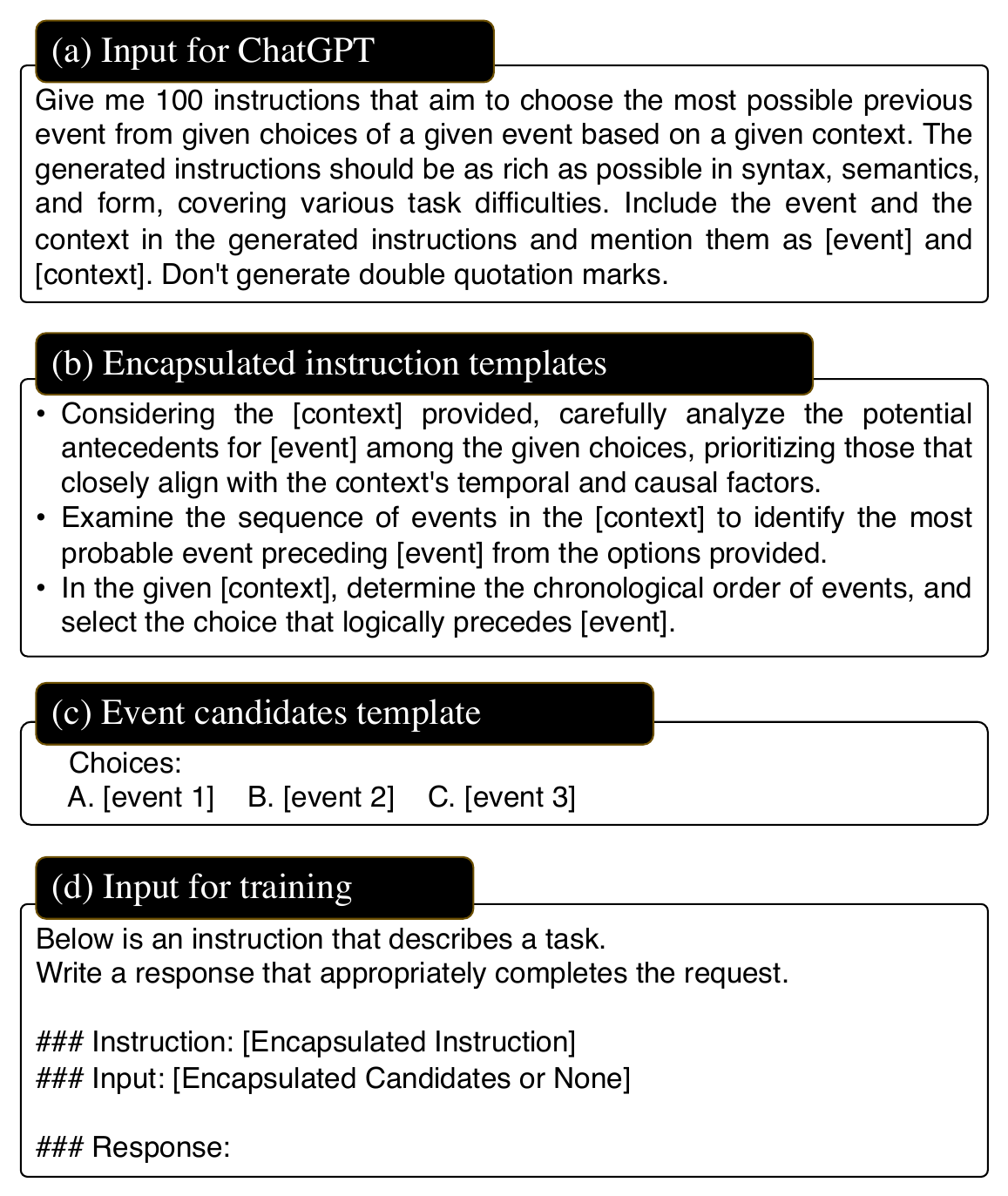}
    \caption{(a) ChatGPT input prompt of \texttt{Before} relation of discrimination learning with context. (b) The ChatGPT generation examples of query in (a). [event] and [context] are placeholders for the head event $\gE^{h}$ and context $\gC$. (c) Template for encapsulating event candidates. (d) The final input for our event-relation training.}
    \label{fig:chatgpt}
\end{figure}
We then query ChatGPT to generate instruction templates. The generation examples are in Figure~\ref{fig:chatgpt} (b). More generated templates are in the Appendix.

After that, we obtain an encapsulated instruction by changing the placeholder [event] by the head event $\gE^{h}$ and the placeholder [context] by the context $\gC$ (if exists). To encapsulate the candidates when in discrimination training, we formulate the choices as a multiple-choice question as shown in Figure~\ref{fig:chatgpt} (c). Based on the acquired encapsulated instruction and event candidates, following Alpaca~\cite{alpaca}, the final inputs are shown in Figure~\ref{fig:chatgpt} (d).

\vspace{-1mm}
\subsection{Counstruction of Event Quadruples}
\label{dc}
In this section, we elaborate on the detail of constructing the \eq. We extract \eqs from BookCorpus~\cite{zhu2015aligning}. 

Initially, we locate tail events which may have associated head events linked by a specific relation. Drawing inspiration from~\citet{zhou2022eventbert}, we identify explicit relation connectives within the PDTB~\cite{prasad2008penn}. For each identified connective, we proceed to locate its child nodes. If any of these child nodes possess a VERB part-of-speech tag, we consider it as the triggering term for the tail event. Subsequently, we traverse the dependency tree originating from the trigger term, capturing a subsection of the tree. Given the sequential nature of the dependency tree, the resultant verb-rooted subsection can be correlated with a span of words, thereby forming a recognized tail event denoted as $\gE^{t}$.

Next, we proceed to extract the head event $\gE^{h}$, relation $\gR$, and the contextual information $\gC$ for \eq. It is important to note that obtaining $\gE^{h}$ is notably more complicated than locating the tail event. This increased complexity arises from the fact that establishing a direct link between the trigger of the head event and the relational connective is often challenging through dependency tree analysis since there may be other nodes intermediately. Rather than relying on linguistic rules for extraction, we employ an end-to-end relation parser similar to the one utilized in ASER~\cite{zhang2020aser}. The function of this relation parser is to dissect a given text where the tail event is. Then extract the head event with a series of relations connecting these two events~\footnote{We only consider $\gE^{h}$ occurring before the tail event.}. The parsed relation is denoted as $\gR$. Within this work, our focus is on the following set of relations:
\begin{align}
        \gR\in\sS^{\gR} = \{&\texttt{Cause}, \texttt{Effect}, \texttt{After}, \\ \notag
    &\texttt{Before}, \texttt{isCond}, \texttt{hasCond}\}.
\end{align}
We only keep relations $\gR\in\sS^{\gR}$. We concatenate sentences before the sentence of $\gE^{h}$ as the context $\gC$. Thus far, we have accomplished the construction of \eqs~$\gQ$. 

We follow~\citet{zhou2022eventbert} to retrieve the negative events to create the candidate event set $\sD$. We build a pool of events from the whole corpus and then retrieve negative events by three heuristic rules. Specifically, given a tail event $\gE^{t}$, we build its negative events, in light of lexicon-based, PoS-based, or in-domain retrieval. Then we sample two events from all negative events and form the candidate event set $\sD$ with the gold tail event $\gE^{t}$.

\label{training}
\section{Experiment}

\begin{table*}[!t]
\centering
\small
\setlength{\tabcolsep}{2.9mm}{\begin{tabular}{lccccccc}

\toprule

 \multirow{2}{*}{$\clubsuit$ \textsc{Close}} &\multicolumn{2}{c}{\textsc{Held-In}} & \multicolumn{2}{c}{\textsc{Held-Out}} &\multicolumn{3}{c}{\textsc{Avg}}\\
\cmidrule(lr){2-3} \cmidrule(lr){4-5}\cmidrule(lr){6-8}

 & \texttt{ECARE}& \texttt{MCTACO}&\texttt{SocialIQA}& \texttt{SCT} & \textsc{Held-In}& \textsc{Held-Out}& \textsc{All}\\
 \midrule
 \multicolumn{8}{c}{\textsc{Large-Scale Models}} \\
 \midrule
 % ChatGPT &0 &78.28&86.72& 5.10~/~0.96~/~6.99~/~5.67&9.68~/~2.91~/~8.46~/~18.91&82.50&&\\
 ChatGPT  &82.36&90.24& 69.68 & 95.88 &86.30&82.78&84.54\\
 Text-Davinci-002 & 76.08 & 90.64 & 73.10 & 95.99 &83.36&84.54&83.95\\
 % text-davinci-003 ~\cite{ouyang2022training}  & 75.38 & 91.45 &&  &83.41&&\\
 \midrule
 \multicolumn{8}{c}{\textsc{7B Models}} \\
  \midrule
   % Flan-T5~\cite{chung2022scaling} &0& 63.27&97.18$^{\star}$&18.45~/~3.62~/~14.63~/~33.33&13.08~/~4.56~/~11.78~/~24.37 &80.22$^{\star}$&&\\
 Alpaca~\cite{alpaca} & 67.73 & 82.49 & 53.43 & 81.77 &75.11&67.60&71.35\\
 Vicuna~\cite{vicuna2023}  & 49.86 & 49.20 & 33.21 & 55.16 &49.53&44.18&46.85\\
 WizardLM~\cite{xu2023wizardlm}  & 54.32 & 68.21 & 34.30 & 53.13 & 61.26 &43.71& 52.48\\
 Dolly-v2~\cite{DatabricksBlog2023DollyV2}  & 49.06 & 44.57 & 33.57 & 49.71 & 46.81&41.64&44.22 \\
 \eis (Ours) &\textbf{77.06}&\textbf{82.80}& \textbf{55.60} & \textbf{87.33} &\textbf{79.93}&\textbf{71.46}&\textbf{75.69}\\

 \bottomrule
\end{tabular}}
\caption{Automatic evaluation results on \textsc{Close} tasks. The metric for \textsc{Close} tasks is accuracy. Bold numbers stand for the best scores of 7B models. } %%% 换了个caption
\label{close} %%% 换了个label
\end{table*}

%%%%%%%%%%%%%%%%%%%%%%%%%%%%%%%%%%%%%%%%%%%%%%%%%%
%%%%%%%%%%%%%%%%%%%%%%%%%%%%%%%%%%%%%%%%%%%%%%%%%%
%%%%%%%%%%%%  第一个表重新画 结束  %%%%%%%%%%%%%%%%%%
%%%%%%%%%%%%%%%%%%%%%%%%%%%%%%%%%%%%%%%%%%%%%%%%%%
%%%%%%%%%%%%%%%%%%%%%%%%%%%%%%%%%%%%%%%%%%%%%%%%%%
% \vspace{-3mm}
\subsection{Evaluation Dataset}
% \vspace{-3mm}
We follow~\citet{tao2023eveval} to incorporate \texttt{ECARE}~\cite{du2022care}, \texttt{MCTACO}~\cite{zhou2019going}, \texttt{SocialIQA}~\cite{sap2019socialiqa}, and \texttt{SCT}~\cite{mostafazadeh2016corpus} to evaluate models' capabilities. These datasets assess the abilities of causal, temporal, intentional event reasoning, and event prediction respectively. For each dataset, we evaluate both \textsc{Close} and \textsc{open} forms of task. In \textsc{Close} form we provide candidates while in \textsc{Open} form we don't. All datasets are the same with \citet{tao2023eveval}. We finally have 8 tasks for test. Note that \texttt{ECARE} and \texttt{MCTACO} are held-in datasets since we explicitly incorporate causal and temporal relations in our event-relational learning. On the contrary, \texttt{SocialIQA} and \texttt{SCT} are held-out tasks.
% \vspace{-2mm}
% \noindent\paragraph{Financial Domain}
% Since there are no event reasoning datasets in the financial domain, we construct datasets for human evaluation. We curated data for \textsc{Open} reasoning in causality, intentional, and event prediction. We create 10 pieces of data for each task in the financial domain.

\vspace{-2mm}

\subsection{Baselines}
We introduce Alpaca-7B~\cite{alpaca}, Vicuna-7B~\cite{vicuna2023}, WizardLM-7B~\cite{xu2023wizardlm}, Dolly-v2-7B~\cite{DatabricksBlog2023DollyV2}, ChatGPT, and InstructGPT~\cite{ouyang2022training} as our baselines. Details are in Appendix A.

\begin{table*}[!t]
\setlength{\belowcaptionskip}{-5mm}
\centering
\small
\setlength{\tabcolsep}{1.2mm}{\begin{tabular}{lccccccc}

\toprule

 \multirow{2}{*}{$\clubsuit$ \textsc{Open}} &\multicolumn{2}{c}{\textsc{Held-In}} & \multicolumn{2}{c}{\textsc{Held-Out}} &\multicolumn{3}{c}{\textsc{Avg}}\\
\cmidrule(lr){2-3} \cmidrule(lr){4-5}\cmidrule(lr){6-8}

 & \texttt{ECARE}& \texttt{MCTACO} &\texttt{SocialIQA}& \texttt{SCT} & \textsc{Held-In}& \textsc{Held-Out} & \textsc{All}\\
 \midrule
 \multicolumn{8}{c}{\textsc{Large-Scale Models}} \\
 \midrule
 % ChatGPT &0 &64.80&89.85& 8.87~/~1.69~/~6.92~/~17.23&6.29/0.86//&&&\\
 ChatGPT & 13.34~/~32.95 & 21.55~/~41.90 & 12.90~/~34.67 & 16.38~/~25.13 & 37.42&29.99&33.70\\
 Text-Davinci-002 & 7.53~/~22.71 & 13.50~/~22.29 &9.00~/~13.79& 12.04~/~19.43 &22.50&16.61&19.55\\
 % text-davinci-003 ~\cite{ouyang2022training}  & 68.41 & 96.04 && 6.79~/~0.82 &&&\\
 \midrule
 \multicolumn{8}{c}{\textsc{7B Models}} \\
  \midrule
   % Flan-T5~\cite{chung2022scaling} &0& 40.07&80.71&10.72~/~5.63~/~12.72~/~17.31&19.53~/~2.60~/~~/~&&&\\
 Alpaca~\cite{alpaca} &10.48~/~17.04& 13.25~/~26.33 & \textbf{7.72}~/~19.48 & \textbf{15.98}~/~25.67&21.68&22.57&22.12\\
 Vicuna~\cite{vicuna2023} & 10.50~/~15.97 & 8.47~/~1.97 & 6.64~/~17.28 & 8.92~/~5.67& 8.97&11.47&10.22\\
 WizardLM~\cite{xu2023wizardlm} &  7.50~/~6.01 & ~~7.85~/~13.66 & 4.31~/~7.45~ & 7.72~/~5.68& 9.83&6.56&8.19 \\
 Dolly-v2~\cite{DatabricksBlog2023DollyV2} & 10.80~/~15.02 & 12.87~/~23.91 & 7.08~/~19.79& 14.64~/~16.52& 19.46&18.15&18.80 \\
 \eis (Ours)&\textbf{10.54}~/~\textbf{28.97}& \textbf{15.60}~/~\textbf{34.93} & 5.12~/~\textbf{27.02} & 13.23~/~\textbf{27.60}&\textbf{31.95}&\textbf{27.31}&\textbf{29.63}\\

 \bottomrule
\end{tabular}}
\caption{Automatic evaluation results on \textsc{Open} tasks in general domain. The metrics for \textsc{Open} tasks are ROUGE-L, and BERT-SCORE. Bold numbers stand for the the best scores of 7B models. \textsc{Avg} for \textsc{Open} task is the average BERT-SCORE.}
\label{open} % 换了个label 
\end{table*}

%%%%%%%%%%%%%%%%%%%%%%%%%%%%%%%%%%%%%%%%%%%%%%%%%%
%%%%%%%%%%%%%%%%%%%%%%%%%%%%%%%%%%%%%%%%%%%%%%%%%%
%%%%%%%%%%%%  第二个表重新画 结束  %%%%%%%%%%%%%%%%%%
%%%%%%%%%%%%%%%%%%%%%%%%%%%%%%%%%%%%%%%%%%%%%%%%%%
%%%%%%%%%%%%%%%%%%%%%%%%%%%%%%%%%%%%%%%%%%%%%%%%%%

\vspace{-1mm}

\subsection{Implementation Settings}
\eis undergoes fine-tuning using academic resources. Precisely, we utilize 4 $\times$ NVIDIA A100 GPUs to train the Llama-7B for 3 epochs. We use the DeepSpeed training framework~\footnote{https://www.deepspeed.ai}, and ZERO-2 strategy along with mixed-precision training (fp16) using the standard AdamW optimizer. The maximum sequence length is set to 512, and the batch size is configured as 32. We use gradient checkpointing. The entire fine-tuning process is completed within a duration of 3 hours.

We use Spacy\footnote{https://spacy.io} for all linguistic extraction. We utilize \eqs instances where both $\gE^{h}$ and $\gE^{t}$ have lengths in 2 to 10 words. We exclude data whose context length falls outside the range of 10 to 50 words. For each \eqs instance, we equally consider training it as either generation or discrimination in event-relational learning. We finally curate 212,538 data for training.

In our pilot experiments, we test multiple input prompts for each model to search for the optimum prompt for evaluation tasks. We observe minimal fluctuations in the results despite prompt variations. To mitigate the impact of other variables, we ensure consistency by employing the same prompt for all models when they undertake the same task. We turn the \textsc{Close} tasks into multiple-choice questions and require the model to answer by the label of choice. All prompts can be found in the Appendix C. 

We find ChatGPT and Vicuna don't generate well-formed events in the zero-shot setting. They generate answers in narrative sentences with explanations leading to difficulty in evaluation. Therefore, we use two-shot in-context learning for them. Other models are in the zero-shot setting.

% \begin{table*}[!t]
% \setlength{\belowcaptionskip}{-3mm}
% % \setlength{\abovecaptionskip}{-1mm}
% \centering
% \small
% \setlength{\tabcolsep}{1mm}{\begin{tabular}{lccccccccc}

% \toprule

%  \multirow{2}{*}{$\clubsuit$ \textsc{Financial}} & \multicolumn{4}{c}{\textsc{Content}}& \multicolumn{4}{c}{\textsc{Format}} &\multirow{2}{*}{\textsc{BS}}\\
%   \cmidrule(lr){2-5}\cmidrule(lr){6-9}
%  & \texttt{Causal}& \texttt{Intentional}&\texttt{Prediction}&\texttt{Avg}& \texttt{Causal}& \texttt{Intentional}&\texttt{Prediction}&\texttt{Avg}&\\
%  \midrule
 
%  Alpaca~\cite{alpaca}&\textbf{4.7}&4.4&\textbf{4.5}&\textbf{4.53}&3.5&3.5&3.5&3.50&26.37\\
% WizardLM~\cite{xu2023wizardlm}&4.3&4.2&3.8&4.10&2.8&3.0&2.3&2.69&16.88\\
% \eis (Ours)&4.6&\textbf{4.5}&4.0&4.36&\textbf{4.7}&\textbf{4.6}&\textbf{4.7}&\textbf{4.66}&\textbf{30.63}\\

% \bottomrule

% \end{tabular}}
% \caption{Evaluation results in the financial domain. Bold numbers stand for the best scores.}
% \label{hf}
% \end{table*}

\vspace{-1mm}
\subsection{Evaluation Metrics}
\vspace{-2mm}
\noindent\paragraph{Automatic Evaluation} We follow~\citet{tao2023eveval} to evaluate all models on automatic metrics. For \textsc{Close} tasks, we use accuracy. In \textsc{Open} tasks, we use ROUGE-L~\cite{lin2004rouge}, and BERT-SCORE~\cite{zhang2019bertscore} metrics for evaluation. 

For \textsc{Close} tasks, some models won't directly generate the label as the answer. We design the following decode protocol to parse the output answers and obtain the final prediction for all models. We show this protocol in the Appendix F.

% \vspace{-2mm}

\noindent\paragraph{Human Evaluation}
One difficulty in automatically evaluating the \textsc{Open} tasks is that the answers for \textsc{Open} tasks may not be unique. Therefore, we also conduct the human evaluation for \textsc{Open} causality, intentional, and prediction tasks. In our evaluation, we focus on two main aspects. Firstly, we assess the content, which involves checking the correctness, reasonableness, and specificity of the generated events. A higher-quality event should accurately align with the queried relation, exhibiting logical coherence and minimal hallucination. Secondly, we examine the format, ensuring that the generated content adheres to the proper structure and completeness expected in an event. We give a score range from 1 to 5 for each aspect and report the average score of the well-educated human evaluators for each data.

\begin{table*}[!t]
\setlength{\belowcaptionskip}{-2mm}
\centering
\small
\setlength{\tabcolsep}{1.5mm}{\begin{tabular}{lcccccccc}

\toprule

 & \multicolumn{4}{c}{\textsc{Content}}& \multicolumn{4}{c}{\textsc{Format}} \\
  \cmidrule(lr){2-5}\cmidrule(lr){6-9}
 & \texttt{Causal}& \texttt{Intentional}&\texttt{Prediction}&\texttt{Avg}& \texttt{Causal}& \texttt{Intentional}&\texttt{Prediction}&\texttt{Avg}\\

\midrule

Alpaca~\cite{alpaca}&3.9&\textbf{3.7}&3.2&3.60&3.2&3.1&3.3&2.86\\
  WizardLM~\cite{xu2023wizardlm}&3.0&3.2&1.8&2.66&2.9&2.4&1.4&2.23\\
  \eis (Ours)&\textbf{4.6}&3.1&\textbf{3.5}&\textbf{3.73}&\textbf{4.7}&\textbf{3.8}&\textbf{3.8}&\textbf{4.10}\\
 \bottomrule

\end{tabular}}
\caption{Human Evaluation results. Bold numbers stand for the best scores.}
\label{hg}
\end{table*}

\begin{table}[!t]
\setlength{\belowcaptionskip}{-5mm}
\centering
\footnotesize
\setlength{\tabcolsep}{1.8mm}{\begin{tabular}{ll}

\toprule
\textsc{Pattern}&\textsc{Example}\\
\midrule
subj-verb-obj&Erika slept part of the trip.\\
subj-verb-prep&Morgan ran down the hallway.\\
subj-verb-xcomp&They want to cast me out.\\
subj-aux-verb-obj&Pierce was taking legal action.\\
subj-verb-ccomp&He smiled that he had survived.\\
subj-verb&A riot of questions surged.\\
subj-verb-obj-prep&I see them through a ripple of smoke.\\
verb-obj&Adopt an outlook on all affairs.\\

\bottomrule

\end{tabular}}
\caption{Top frequent event patterns. }
\label{pattern}
\end{table}

\subsection{Results}
\vspace{-3mm}
\noindent\paragraph{\textsc{Close} Tasks} We show evaluation results of \textbf{Close} tasks in Table~\ref{close}. We first find \eis performs well in \textsc{Held-In} tasks. \eis outperforms all other instruction-tuning models both in \textsc{Held-In} and \textsc{Held-Out}. \eis obtains 75.69 overall average \textsc{Close} score which is 4.34 higher than the second best Alpaca. In \texttt{ECARE} dataset, \eis event achieves better results than Text-Davinci-002. The results demonstrate the effectiveness of our event-oriented instruction tuning. \eis can better associate event knowledge to distinguish the correct event from event candidates.

We also find \eis performs well in \textsc{Held-Out} tasks. \eis outperforms all other instruction-tuning models both in \texttt{SocialIQA} and \texttt{SCT} and obtains a 71.46 average score which is 3.86 higher than the second-best Alpaca. The results demonstrate that \eis can transfer event knowledge to other event reasoning tasks or event relations.

\vspace{-1mm}
\noindent\paragraph{\textsc{Open} Tasks} We report automatic evaluation of \textsc{Open} Tasks in Table~\ref{open}. We find \eis performs well in ROUGE-L and Bert-Score. The average \textsc{Open} Bert-Score of our model is 7.51 higher than the second-best Alpaca. This result shows that \eis can understand the event semantics more and generate better structures and semantics.

% In all, \eis is competitive in causality and temporal event reasonings. The average of \textsc{Close} and \textsc{Open} is 7.55 than the second best instruction-tuning model. Our event-oriented training paradigm can enhance the comprehension of the structure and semantics of events. 

% \noindent\paragraph{Held-Out Tasks} The automatic results of held-out tasks are shown in Table~\ref{heldout}. We first find \eis performs well in held-out \textsc{Close} tasks. \eis outperforms all other instruction-tuning models both in \texttt{SocialIQA} and \texttt{SCT}. \eis obtains 71.46 average \textsc{Close} score which is 3.86 higher than the second best Alpaca. The results demonstrate that \eis can transfer event knowledge to other event reasoning tasks or event relations.

% Similar to held-in \textsc{Open} tasks, we find \eis performs better in ROUGE-L and Bert-Score while worse in BLEU-1 and BLEU-2. This consistent phenomenon demonstrates \eis comprehends event semantics and can transfer this ability to other unseen tasks.

\begin{table*}[!t]
\setlength{\belowcaptionskip}{-5mm}
\centering
\footnotesize
\setlength{\tabcolsep}{1mm}{\begin{tabular}{p{4cm}p{1.5cm}p{1.5cm}p{4cm}p{4cm}}

\toprule
% &$\gC$&$\gE^{h}$&$\gR$&$\gE^{t}$ \\
\textsc{Question}& \textsc{Label}&\ei& \textsc{Alpaca}&\textsc{WizardLM}\\
\midrule
% \multicolumn{5}{c}{\textsc{General}}\\
% \midrule

Context: Sasha gave the dog some pills to help it feel better.Question: What will Sasha want to do? Answer:&Hope that it gets well. &She could take him to the vet.  &Sasha will likely want to monitor the dog's progress to make sure the pills are working and that the dog is feeling better. &Based on the context, it is not clear what Sasha wants to do next. Can you provide more information or context?\\

\midrule
What is the cause of the "Lysosomes have less acidic property than the cytosol."? Answer:& A higher ph indicates less acidic properties. & They have a higher pH. 
& Lysosomes are membrane-bound organelles that contain enzymes that are able to break down cell waste products.&The statement that describes the difference in pH levels between the lysosomes and the cytosol of a cell. \\

\bottomrule

\end{tabular}}
\caption{Generation cases of different models on \textsc{Open} tasks. }
\label{case}
\end{table*}

% \vspace{-1mm}
\noindent\paragraph{Human Evaluation} We conduct a human evaluation of three \textsc{Open} tasks. We assess \textsc{Content} and \textsc{Format} aspects for all tasks. We find this human evaluation is consistent with automatic evaluation. \eis achieves highest scores in both \textsc{Content} and \textsc{Format}. These results further demonstrate the effectiveness of our model. Our model can answer the event relational reasoning tasks in a way that human favors more. It can generate more precisely and concisely. The generations are more readable and understandable by humans. The events generated are more complete than others. The results also indicate that \eis can generate more confidently without extra guesses by generally trained models. We find, in the \texttt{intentional} task, \eis falls behind Alpaca in \textsc{Content}. This result may be due to the training relation we choose.  Since \texttt{intentional} is a held-out relation, there may exist a misalignment of generations of intention content. Overall, \eis performs the best under human evaluation on average.
% \eis performs worse on it than other tasks.

% \subsection{Financial Domain}

% We evaluate the performance of models in the financial domain. The automatic metric Bert-Score shows the effectiveness of our model in the financial domain. The Bert-Score of \eis is 4.25 higher than the second-best Alpaca. In our human evaluation, we find \eis excels other models in \textsc{Foirmat} aspect. This shows the generation of our model is complete in the event. Moreover, our model can answer in a more clear and precise way while other models would return lengthy and irrelevant words and sentences. We find \eis falls behind Alpaca on content average. This is probably attributed to that our training data is curated from BookCorpus where financial content is rare. Changing the raw corpus or mixing it with financial corpus would mitigate this shortage. In all, the event comprehension ability of our model can transfer across domains. 

\subsection{Case Analysis}

\vspace{-1mm}

\noindent\paragraph{Event Structure} 
We show the top frequent event structure patterns in Table~\ref{pattern}. We obtain the pattern by extracting the root verb and its direct children of an event according to dependency parsing results. We find our \eqs maintains the completeness of events and covers stereotypical patterns. We also show the length distribution of events in the Appendix C. We notice the events are diversified in patterns and lengths. 

\noindent\paragraph{Event Distribution}
In Figure~\ref{fig:verb}, we showcase a word cloud of verbs of \eq. We find our curated \eqs covers a spectrum of event types. This is the main reason that \eis is able to integrate event knowledge and reason events in various domains.

% \vspace{-2mm}
\noindent\paragraph{Evaluation Cases} We showcase several cases of \textsc{Open} task generations of three models in Table~\ref{case}. In the first example, \eis can generate precise cause intent of the head event. The generation is also as concise as the label. Alpaca could generate the correct intent of the subject, however, the generation is lengthy. The WizardLM fails to output the answer. In the second example, \eis also answers correctly about the cause of the head event. Alpaca and WizardLM make predictions with excessive association and hallucination. These cases further demonstrate that \eis can associate correct event knowledge and maintain the completeness of generated events. Furthermore, \eis can make concise inferences among all models.

\begin{figure}[!tb]
\setlength{\belowcaptionskip}{-5mm}
    \centering
    \includegraphics[width=0.8\columnwidth]{latex/figs/words_500.pdf}
    \caption{Wordcloud of verbs of events. }
    \label{fig:verb}
\end{figure}
\vspace{-1mm}

\section{Related Work}
\vspace{-2mm}
\noindent\paragraph{Event Reasoning}
Event relational reasoning infers events of certain inter-relations. \citet{du2022care} aims to select the accurate cause or effect event from candidates. \citet{zhou2019going} serves as a dataset for event temporal reasoning. Current works presented a scenario for current language understanding and generation systems by incorporating the need for counterfactual reasoning~\cite{qin2019counterfactual, qin2020back}.
In addition to single-event relation reasoning, existing works also reason events according to diversified event relations~\cite{poria2021recognizing, han2021ester, yang2022towards}. \citet{tao2023unievent} further unifies datasets of several event-inter relations to transfer event relational knowledge to unseen tasks. 

Predicting events necessitates the model to anticipate forthcoming occurrences grounded in the present context ~\cite{zhao2021event}. \citet{mostafazadeh2016corpus} employs a multiple-choice framework to predict future events by encompassing a diverse range of common-sense connections among events. \citet{guan2019story} establish a dataset oriented towards capturing event logic, enabling the generative prediction of future incidents.

% Event prediction requires the model to forecast future events based on current circumstances~\cite{zhao2021event}. \citet{mostafazadeh2016corpus} anticipates forthcoming events involving a wide array of common-sense relations among events through a multiple-choice structure. On the other hand, \citet{guan2019story} introduces a dataset aimed at capturing the logic behind events to predict future occurrences in a generative manner.

\citet{tao2023eveval} present the Event Semantic Processing including the event understanding, reasoning, and prediction of event semantics.

\vspace{-2mm}
\noindent\paragraph{Instruction Tuning}
Instruction tuning refers to the process of fine-tuning a large language model based on specific instructions or guidance provided during training. \citet{chung2022scaling} finetunes on T5 with a scaling number of datasets which achieves strong few-shot performance even compared to much larger models. \citet{alpaca} is trained by fine-tuning the LLaMA~\cite{touvron2023llama} model using a dataset consisting instructions generated by text-davinci-003. \citet{vicuna2023} is an open-source chatbot created by fine-tuning LLaMA using user-shared conversations gathered from ShareGPT. \citet{xu2023wizardlm} extends the previous model by evolve-instruct algorithms to improve the model. \citet{DatabricksBlog2023DollyV2} leverages data on the Databricks platform. 

In another line of research, instruction tuning is used to make a language model more focused and specialized in certain abilities or domains. \citet{zhang2023huatuogpt} trains a medical conversation model with different sources of datasets with instructions. 
% \citet{fan2023grammargpt} constructed about 1k parallel data and utilized these data to fine-tune open-source LLMs.
\citet{cui2023chatlaw} propose a legal LLM named ChatLaw by legal domain dataset and mitigate hallucination of the model.
\citet{deepke-llm} train an LLM specialized for information extraction with data adapted from a knowledge graph.
\citet{yang2023fingpt} design an automatic data curation pipeline and in building financial open-source LLM.
\citet{tang2023toolalpaca} propose a dataset to improve the tool manipulating ability of LLMs.
Our work lies in this ability enhancement line of research.

\vspace{-2mm}

\noindent\paragraph{Event-Aware Pretraining}
Considering both the pre-training and fine-tuning strategies, researchers are dedicated to improving event processing through fine-tuning techniques that incorporate events. In their study, \citet{yu2020cocolm} inject intricate commonsense knowledge about events into pre-trained language models. Similarly, \citet{zhou2022eventbert, zhou2022claret} enhance language models by focusing on event-related tasks through event masking prediction and generation. However, these works struggle to effectively perform zero-shot reasoning.
\section{Conclusion}
In this study, we introduce Event-Oriented Instruction Tuning to enhance event reasoning capabilities and train our model \ei. 
We first introduce a novel structure called \eqs as a foundational structure. Building upon this, we establish event relation learning through instruction tuning using generated prompts. We create an instruction-tuning dataset focused on events, encompassing comprehensive and diversified event data both in syntax and semantics.
% Our \eqs structure is constructed using heuristic techniques. 
Subsequently, we fine-tune Llama to create the \eis model. We conduct experiments on both \textsc{Close} and \textsc{Open} task settings and compare with several strong cutting-edge instruction-tuned models. Through extensive experiments on 8 datasets, the outcomes demonstrate the efficacy of our proposed approach.

\section*{Limitations}
In this paper, we only achieve a model that excels in textual event reasoning. However, the event can be represented in other modalities such as visual data. Images would contain more information beyond sentences of events. Leveraging data from other modalities to improve performance remains challenging. We leave it to future work.

\bibliography{main}

\appendix

\newpage

\section{Decoding Protocol}
We show our decoding protocol for extracting answers of \textsc{Close} tasks as follows:

\begin{algorithm}
\small
\SetAlgoNoEnd
\SetKwInOut{Input}{Input}
\SetKwInOut{Output}{Output}
\SetKw{KwTo}{in}
\SetKwIF{If}{ElseIf}{Else}{if}{then}{else if}{else}{endif}

\Input{Prediction $\gP$, candidate set $\sD$.}
\Output{Answer $\gA$.}

pattern = \text{"the(?: correct)? (?:option$|$answer) is[}\textbackslash\text{ s:]+([A-H])"} \\
\If{$\gP$$\mathrm{.startsWithAlphabet()}$}{
    $\gA$ = starts\_alphabet
}
\uElseIf{
$\mathrm{re.match(pattern}$, $\gP$)
}
{
    $\gA$ = $\mathrm{re.extract(\gP, patten)}$
    % Extract the \textit{Answer} follow the \textit{pattern} from \textit{Prediction}.
}
\Else{
    $\gA$=$\underset{c \in \sD}{\mathrm{argmax}}$($\mathrm{WordOverlap}$($c$, $\gP$))
}
\KwRet{$\gA$}
% \caption{\textsc{Close} answer decoding.}
\label{decode}

\end{algorithm}

% \begin{algorithm}
% \begin{algorithmic}
% \State \textit{pattern} = "the(?: correct)? (?:option$|$answer) should be[\textbackslash s:]+([ABCDEFGH])" \\
% \If {\textit{Output} starts with an alphabetical number}
%     \State Set \textit{prediction} as the alphabetical number
% \ElsIf {re.match(\textit{pattern}, \textit{Output})}
%     \State Extract the \textit{prediction} follow the \textit{pattern}.
% \Else 
%     \State \textit{prediction}=$\underset{c \in \sD}{argmax}$($\mathrm{WordOverlap}$($c$, \textit{Ouput})

% \EndIf\algnotext{endif}
% \end{algorithmic}
% \end{algorithm}

\begin{figure}[!h]
    \centering
    \includegraphics[width=1\columnwidth]{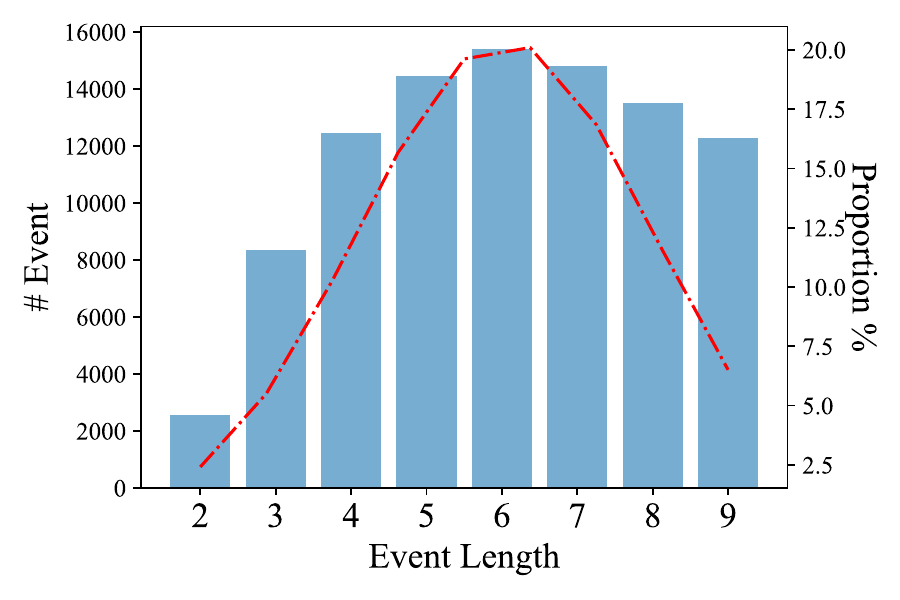}
    \caption{Statistic of the length of events. }
    \label{fig:len}
\end{figure}

\section{Baselines}

\noindent\paragraph{Alpaca} Vicuna, an open-source chatbot, is developed by fine-tuning LLaMA using user-shared conversations collected from ShareGPT. Preliminary evaluations with GPT-4 as the evaluator reveal that Vicuna achieves more than 90\% quality when compared to ChatGPT.
\noindent\paragraph{Vicuna}This particular model undergoes training through fine-tuning the LLaMA 7B model with a dataset containing 52,000 demonstrations accompanied by instructions generated using Text-Davinci-003.
\noindent\paragraph{WizardLM} WizardLM is trained on instruction-tuning data generated by the Evol-Instruct algorithm. It demonstrates remarkable performance on complex tasks and remains competitive across various metrics.
\noindent\paragraph{Dolly-v2} Databricks' Dolly-v2 7B is a sizable language model designed for instruction-following, trained using 15,000 instruction/response fine-tuning records created by Databricks employees. These records cover various capability domains, encompassing classification, closed QA, generation, information extraction, open QA, and summarization.
\noindent\paragraph{ChatGPT} An extensive language model developed by OpenAI\footnote{https://chat.openai.com/}. The model undergoes fine-tuning, employing a combination of supervised and reinforcement learning techniques to enhance its performance.
\noindent\paragraph{InstructGPT} We assess two InstructGPT models, specifically Text-Davinci-002.

% \newpage
\section{Event Length}
We show the length distribution of events in Figure~\ref{fig:len}.

\section{Event Reasoning Evaluation Prompts}
\label{erp}
\begin{figure*}[!th]
    \centering
    \includegraphics[width=2\columnwidth]{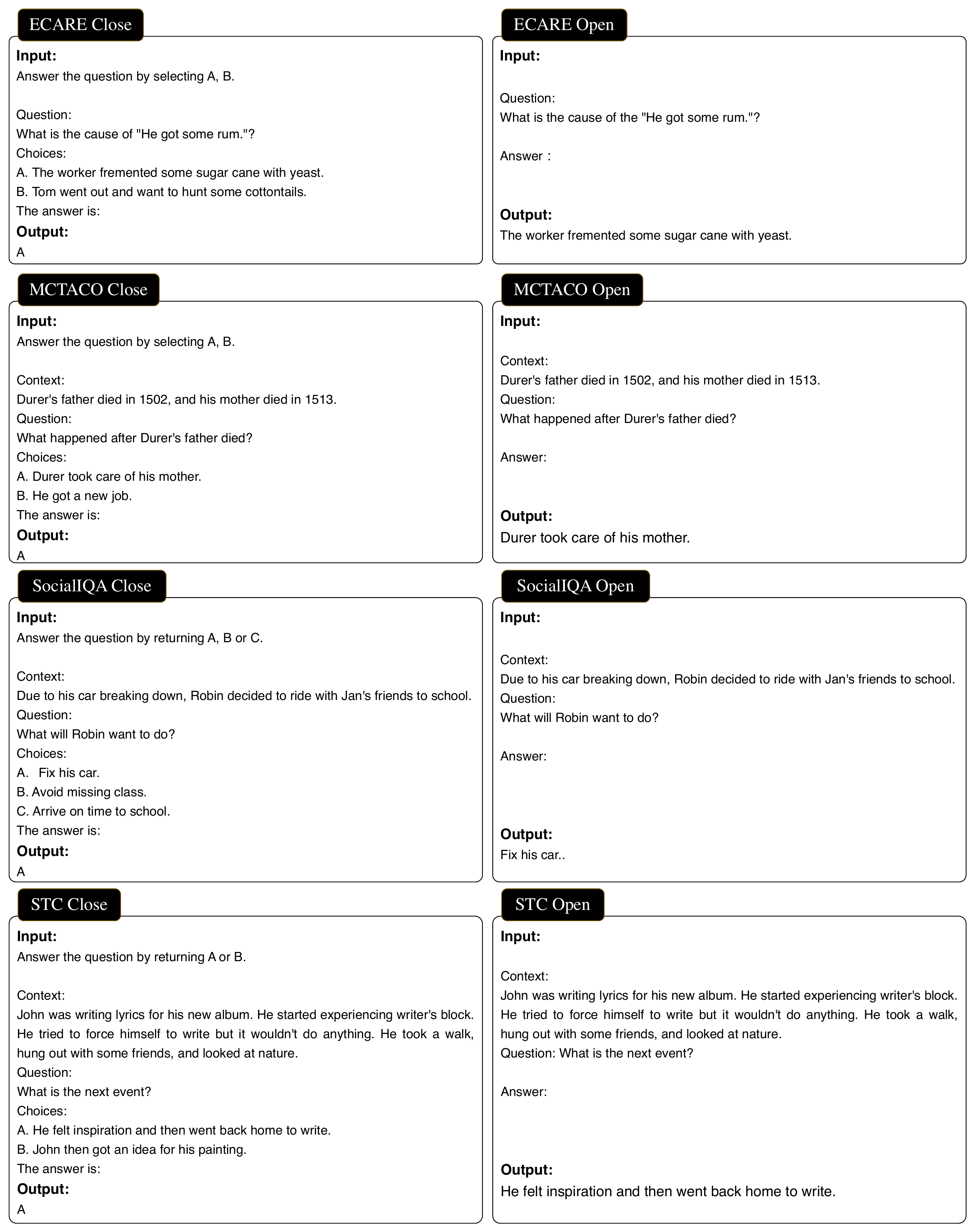}
    \caption{Evaluation prompts for all models.}
    \label{fig:gp}
\end{figure*}

We show prompts for evaluation on all tasks for all models in Figure~\ref{fig:gp}. 
\section{Input for ChatGPT}
\label{ic}
\begin{figure*}[!th]
\setlength{\belowcaptionskip}{10mm}
    \centering
    \includegraphics[width=2.1\columnwidth]{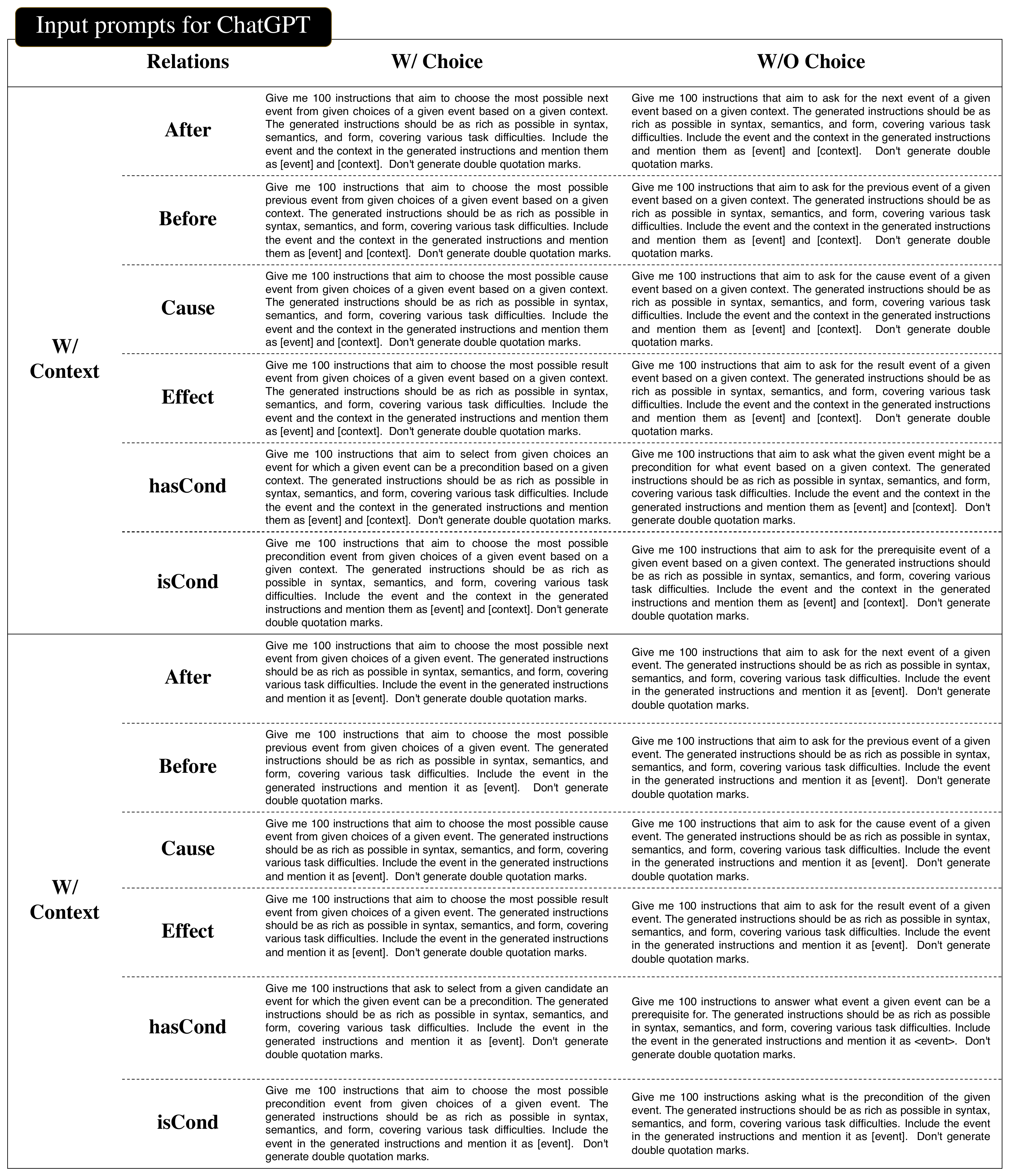}
    \caption{Input for ChatGPT to generate instruction-tuning templates.}
    \label{fig:ic}
\end{figure*}

We show ChatGPT input for generating instruction templates in Figure~\ref{fig:ic}.
\section{Examples of Instruction Templates}
\label{ip}
\begin{figure*}[!th]
    \centering
    \includegraphics[width=2.1\columnwidth]{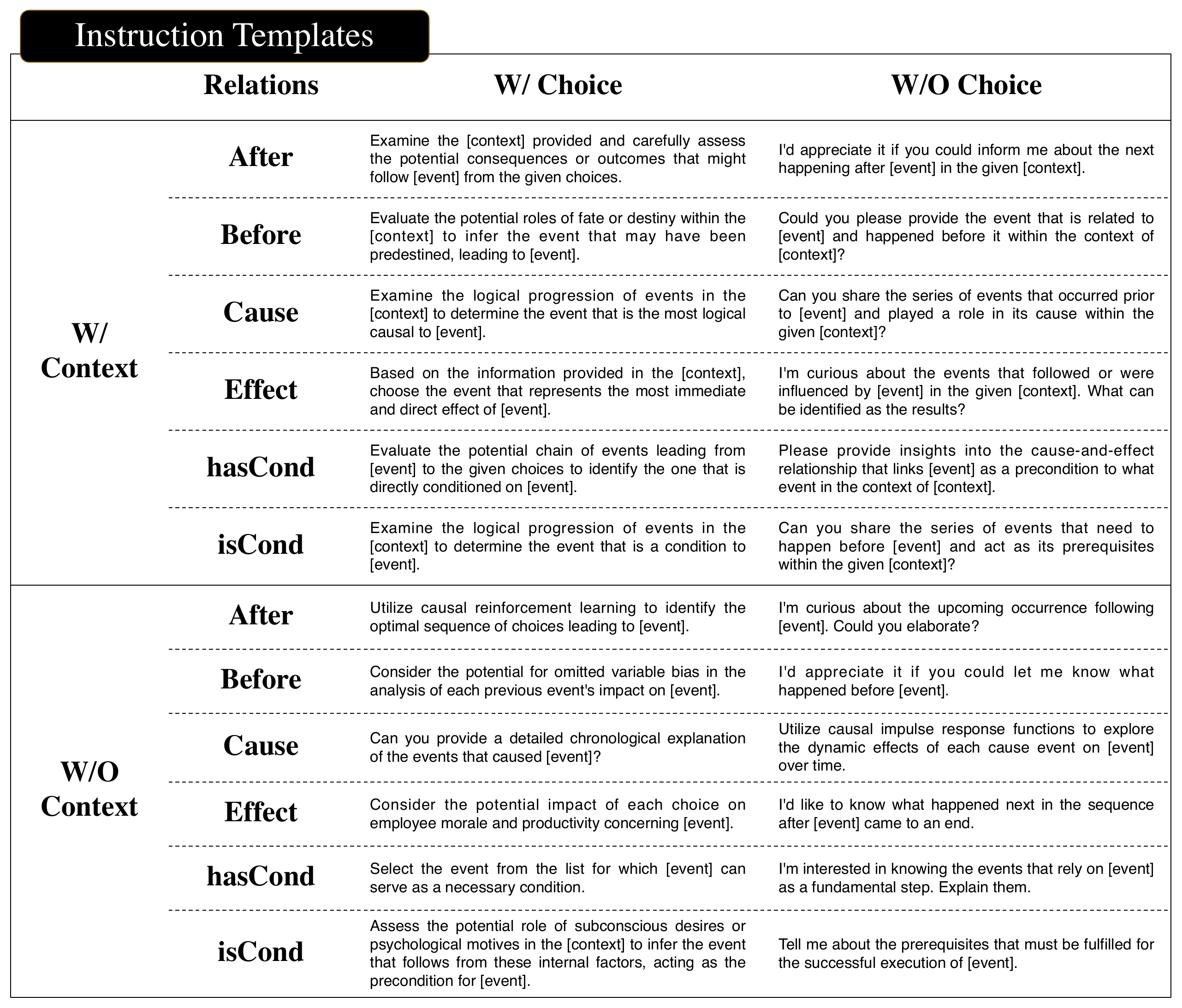}
    \caption{Examples of instruction templates generated by ChatGPT.}
    \label{fig:ip}
\end{figure*}

We showcase examples of instruction templates in Figure~\ref{fig:ip}.

% \newpage

% \begin{figure*}[!th]
% \setlength{\belowcaptionskip}{40mm}
% % \setlength{\abovecaptionskip}{-3mm}
%     \centering
%     \includegraphics[width=2.1\columnwidth]{latex/figs/financial_prompt.pdf}
%     \caption{Evaluation prompts for all models in the financial domain.}
%     \label{fig:fp}
% \end{figure*}

\end{document}